\documentclass[sigconf, nonacm]{acmart}

\newcommand\vldbdoi{XX.XX/XXX.XX}
\newcommand\vldbpages{XXX-XXX}
\newcommand\vldbvolume{14}
\newcommand\vldbissue{1}
\newcommand\vldbyear{2020}
\newcommand\vldbauthors{\authors}
\newcommand\vldbtitle{\shorttitle} 
\newcommand\vldbavailabilityurl{https://beaverbench.github.io/}
\newcommand\vldbpagestyle{plain} 

\usepackage{graphicx}
\usepackage{adjustbox}
\usepackage{makecell}
\usepackage{booktabs}
\usepackage{listings}
\usepackage{cleveref}
\usepackage{float}
\usepackage{tcolorbox}
\usepackage{xspace}
\usepackage{tikz}
\usepackage{pifont}

\begin{document}
\title{BEAVER: An Enterprise Benchmark for Text-to-SQL}

\author{Peter Baile Chen}
\affiliation{\institution{MIT}}
\email{peterbc@mit.edu}

\author{Devin Yang}
\affiliation{\institution{MIT}}
\email{kevin023@mit.edu}

\author{Weiyue Li}
\affiliation{\institution{Harvard University}}
\email{weiyueli@fas.harvard.edu}

\author{Fabian Wenz}
\affiliation{\institution{TU Munich and MIT}}
\email{fab_wenz@mit.edu}

\author{Yi Zhang}
\affiliation{\institution{Greenshoe, Inc.}}
\email{yi@greenshoe.ai}

\author{Nesime Tatbul}
\affiliation{\institution{Intel and MIT}}
\email{tatbul@csail.mit.edu}

\author{Michael Cafarella}
\affiliation{\institution{MIT}}
\email{michjc@csail.mit.edu}

\author{Çağatay Demiralp}
\affiliation{\institution{AWS AI and MIT}}
\email{cagatay@csail.mit.edu}

\author{Michael Stonebraker}
\affiliation{\institution{MIT}}
\email{stonebraker@csail.mit.edu}


\begin{abstract}
Existing text-to-SQL benchmarks have largely been constructed from public databases with well-structured schemas and simplistic question-SQL pairs. While large language models (LLMs) excel on these settings, their efficacy in complex private enterprise environments, characterized by intricate schemas, domain knowledge, and analytical user queries involving sophisticated structures and functions, remains unproven. To bridge this gap, we introduce \dataset{}, the first text-to-SQL benchmark derived from \textit{private data warehouses}. It comprises 9128 question-SQL pairs sourced from real-world query logs and 812 tables across 19 diverse domains.
Building this benchmark is challenging because (1) enterprise query logs are scarce due to privacy constraints, and (2) existing all-or-nothing evaluation metrics based on accuracy make error diagnosis difficult---especially when producing a correct query involves solving multiple compounded challenges, such as domain knowledge and query complexity.
We address these issues at two levels. At the dataset level, we synthesize high-fidelity, expert-verified queries that increase dataset size and isolate individual challenges or combine them, producing queries focused on domain knowledge, query complexity, and both.
At the evaluation level, we provide human annotations and evaluation metrics for five \textit{subtasks} critical to successful query generation to enable fine-grained analysis: multi-table retrieval, join key detection, column mapping, domain knowledge extraction, and query decomposition.
Our evaluation reveals a significant performance gap compared to existing benchmarks: state-of-the-art agentic frameworks using the advanced model \texttt{GPT-5.2} achieve only 10.8\% accuracy.
When provided with all subtask annotations as oracle hints, accuracy increases to 30.1\%, confirming that a major bottleneck lies in correctly resolving these subtasks.
Finally, we provide a taxonomy of the residual errors that persist even with subtask hints, identifying specific challenges such as the use of advanced functions. We believe \dataset{} will serve as a critical benchmark for developing the next generation of robust, enterprise-ready SQL agents.



\end{abstract}


\newcommand{\dataset}{BEAVER}
\newcommand{\reforce}{ReFoRCE}
\newcommand{\circled}[1]{%
  \begin{tikzpicture}[baseline=(char.base)]
    \node[shape=circle, draw, inner sep=0.8pt, fill=black, text=white, font=\bfseries\small] (char) {#1};
  \end{tikzpicture}%
}

\maketitle

\pagestyle{\vldbpagestyle}
\begingroup\small\noindent\raggedright\textbf{PVLDB Reference Format:}\\
\vldbauthors. \vldbtitle. PVLDB, \vldbvolume(\vldbissue): \vldbpages, \vldbyear.\\
\href{https://doi.org/\vldbdoi}{doi:\vldbdoi}
\endgroup
\begingroup
\renewcommand\thefootnote{}\footnote{\noindent
This work is licensed under the Creative Commons BY-NC-ND 4.0 International License. Visit \url{https://creativecommons.org/licenses/by-nc-nd/4.0/} to view a copy of this license. For any use beyond those covered by this license, obtain permission by emailing \href{mailto:info@vldb.org}{info@vldb.org}. Copyright is held by the owner/author(s). Publication rights licensed to the VLDB Endowment. \\
\raggedright Proceedings of the VLDB Endowment, Vol. \vldbvolume, No. \vldbissue\ %
ISSN 2150-8097. \\
\href{https://doi.org/\vldbdoi}{doi:\vldbdoi} \\
}\addtocounter{footnote}{-1}\endgroup

\ifdefempty{\vldbavailabilityurl}{}{
\vspace{.3cm}
\begingroup\small\noindent\raggedright\textbf{PVLDB Artifact Availability:}\\ 
The source code, data, and/or other artifacts have been made available at \url{\vldbavailabilityurl}.
\endgroup
}

\begin{figure*}
\centering
\includegraphics[page=1, width=\linewidth, clip]{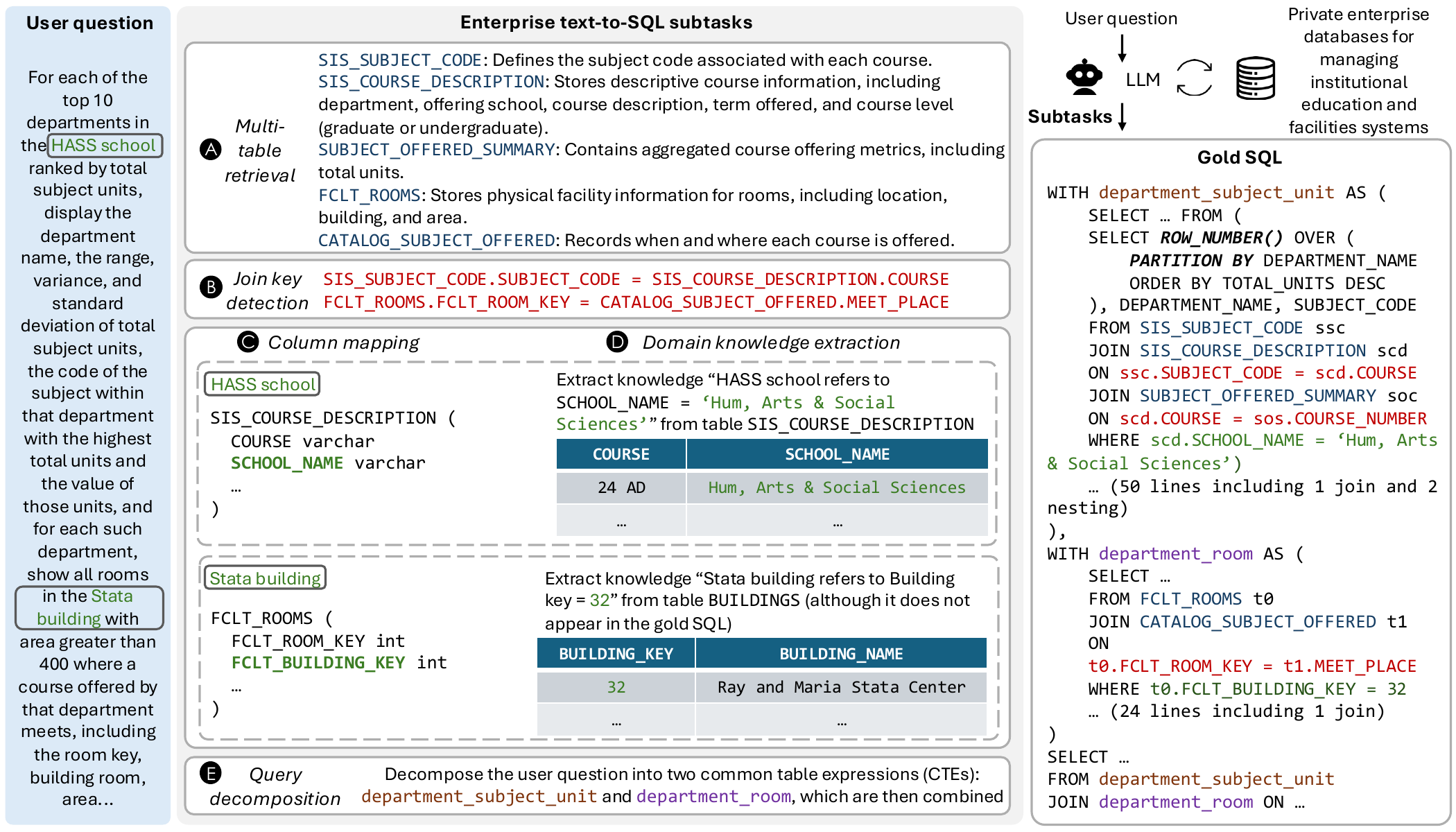}
\caption{A representative enterprise text-to-SQL example on private institutional databases for education and facilities management, highlighting the five key subtasks critical to success.
The LLM should address subtasks when constructing the SQL statement.
Due to space constraints, we only show an abbreviated portion of the gold SQL, which spans more than one hundred lines, along with a few examples of join key detection, column mapping, and domain knowledge extraction.
}
\label{fig:example}
\end{figure*}

\section{Introduction}
\label{sec:intro}



Tables are a fundamental medium for storing and analyzing information. 
Traditionally, querying tables requires writing SQL statements, which can be challenging for non-technical users and time-consuming even for experts when queries involve complex business logic. Recent advances in large language models (LLMs) have significantly improved the feasibility of automatically translating natural language questions into SQL, or text-to-SQL. On popular benchmarks such as Spider~\citep{yu2018spider} and BIRD~\citep{li2024can}, LLMs achieve strong performance. For example, \texttt{GPT-4o} attains approximately 82\% accuracy on the original BIRD dataset~\citep{shkapenyuk2025automatic}. On a corrected version of BIRD with annotation errors fixed~\citep{Jin2026TexttoSQLBA}, \texttt{GPT-4o} combined with ChessSQL~\citep{Talaei2024CHESSCH} still achieves around 81\% accuracy.

Despite this success, existing benchmarks largely reflect a simplified setting. They are constructed from publicly available databases with relatively small numbers of tables, clean and descriptive schema, explicit or easily inferred join relationships, and question-SQL pairs that typically involve only a limited number of joins or shallow nesting. As a result, the strong performance of current models and methods on existing datasets may not carry over to real-world private enterprise environments, where both the data and the user queries are substantially more complex.

Private enterprise data warehouses are designed for internal business operations rather than external accessibility. Their schemas are often large, heterogeneous, and continuously evolving, containing hundreds or thousands of tables with cryptic column names and limited documentation. Join relationships are frequently implicit rather than explicitly declared. Successfully answering a user question in this setting involves retrieving the correct subset of tables from a vast schema, identifying joinable columns without prior hints, mapping natural language concepts to obscure column names, and extracting domain-specific knowledge that is not explicitly encoded in the schema. These challenges are further compounded by the fact that the underlying data is private: public LLMs cannot have prior exposure to the domain-specific entities, codes, or conventions used in such databases, a setting in which LLMs are known to struggle~\citep{kandpal2023large}.

To concretely illustrate these challenges, \Cref{fig:example} presents a representative enterprise query from private institutional databases for education and facilities management. Answering the user question involves
\protect\circled{A} retrieving five relevant tables from a large database and
\protect\circled{B} inferring implicit join relationships, such as joining \\
\texttt{FCLT\_ROOMS.FCLT\_ROOM\_KEY} with \\
\texttt{CATALOG\_SUBJECT\_OFFERED.MEET\_PLACE}, even though no foreign key constraints are defined and the column names differ significantly.
\protect\circled{C} Moreover, the phrase ``all rooms in Stata building'' cannot be resolved directly from the \texttt{FCLT\_ROOMS} table. Instead, the model must identify the column \texttt{FCLT\_ROOMS.FCLT\_BUILDING\_KEY} and \protect\circled{D} consult the \texttt{BUILDINGS} table, even though it does not appear in the final SQL statement, to determine that ``Stata building'' corresponds to the internal building key \texttt{32}. This column mapping and domain knowledge must then be expressed as the predicate\\
\texttt{FCLT\_ROOMS.FCLT\_BUILDING\_KEY = 32}.

Beyond schema complexity, enterprise user queries are themselves substantially more sophisticated. Instead of simple lookups, they are typically analytical, involving advanced SQL constructs such as multiple joins, nested subqueries, Common Table Expressions (CTEs), and window or scalar functions.
As illustrated in \Cref{fig:example}, the gold SQL for the example question exceeds 100 lines and includes six joins and three levels of nesting. \protect\circled{E} The query involves decomposition and is structured using two CTEs, one for computing department subject units and another for identifying department rooms, which are later combined to produce the final result. Additionally, answering the requirement of ``top 10 departments ranked by total subject units'' necessitates advanced window functions, including \texttt{ROW\_NUMBER()} and \texttt{PARTITION BY}.

To rigorously study text-to-SQL in this underexplored but practically critical setting, we introduce \dataset{}, the first text-to-SQL benchmark derived from real \textit{private enterprise data warehouses}. It is built from three anonymized enterprise warehouses, with SQL statements sourced from actual user query logs and production reports, and corresponding natural language questions formulated in collaboration with experienced database administrators.

Constructing such a benchmark, covering both dataset creation and evaluation, poses two key issues.
First, at the dataset level, privacy and security constraints limit access to large volumes of real enterprise query logs.
Second, at the evaluation level, the standard execution accuracy metric (1 if equal, 0 otherwise) based on LLM-generated and gold SQL outputs~\citep{yu2018spider,li2024can} is inherently all-or-nothing. Yet, answering enterprise queries often involves simultaneously resolving multiple challenges, such as domain knowledge and query complexity, to produce fully correct results. Therefore, existing metrics make it difficult to measure performance and progress along individual challenge dimensions.
We tackle these issues at two levels.

At the dataset level, we propose a systematic data synthesis pipeline, \textit{Structural Template Recomposition}. The pipeline extracts high-quality, representative templates capturing base\\
\texttt{SELECT-FROM-JOIN-WHERE-AGGREGATION} structures, nesting, CTE usage, and domain knowledge from real queries. These templates are iteratively composed with different tables, columns, and domain-specific instances to produce high-fidelity queries. By selectively combining templates, we generate queries that isolate individual challenges: queries with domain knowledge but low complexity, queries with high complexity but no domain knowledge, and queries combining both domain knowledge and high complexity. Experts verify all synthesized queries to ensure executability and correctness. This pipeline increases dataset scale while enabling more balanced and fine-grained diagnostic evaluation across individual challenge dimensions.

At the evaluation level, we go beyond the all‑or‑nothing approach of execution accuracy metric by introducing fine‑grained evaluation across five subtasks critical to successful query generation: multi‑table retrieval, join key detection, column mapping, domain knowledge extraction, and query decomposition. To support this analysis, each query is annotated not only with the gold question-SQL pair but also with human‑annotated labels for all subtasks, along with their corresponding evaluation metrics.

Overall, \dataset{} contains 9128 queries and features large\\ enterprise‑scale schemas, averaging 101.5 tables and 869.4 columns per database across 19 diverse domains such as facilities, networking, and education. On average, queries contain 316.7 tokens and frequently involve multiple tables, numerous joins, advanced functions, nested structures, and Common Table Expressions (CTEs). For comparison, BIRD~\cite{li2024can} databases contain only 6.8 tables and 72.5 columns on average, with queries spanning 37.3 tokens. Even the recently introduced Spider~2.0 benchmark~\cite{lei2025spider2} (already regarded as significantly more challenging than BIRD) includes only 52.6 tables and 803.6 columns per database, with queries averaging 144.5 tokens (see \Cref{tab:stats} for details).

We conduct extensive evaluations across seven recent LLMs and four text-to-SQL methods on \dataset{} and observe a substantial performance gap compared to existing benchmarks. Even advanced agentic methods using the advanced model \texttt{GPT-5.2} struggle on subtasks and achieve only 10.8\% execution accuracy.
When provided with all subtask annotations as oracle hints, execution accuracy improves dramatically to 30.1\%, indicating that these subtasks represent key bottlenecks.
Finally, we perform a comprehensive error analysis of the residual failures, highlighting difficulties with analytical SQL constructs (e.g., grouping, ordering), advanced SQL functions (e.g., window and scalar functions), and additional errors that persist despite oracle guidance.

In summary, our contributions are fourfold.
(1) We introduce \dataset{}, the first benchmark for text-to-SQL in private enterprise settings, featuring real schemas and queries.
(2) We propose Structural Template Recomposition, a principled approach to expand enterprise datasets under strict privacy constraints and enable fine-grained evaluation along individual challenge dimensions.
(3) We conduct a comprehensive evaluation of state-of-the-art methods, revealing their key limitations. To support fine‑grained analysis beyond coarse execution accuracy, we introduce five critical subtasks along with their evaluation metrics.
(4) We present a detailed taxonomy of residual errors that highlights directions for building more robust, enterprise-ready SQL agents.

The remainder of the paper is organized as follows.
\Cref{sec:related} reviews related work.
\Cref{sec:subtasks} details the subtasks of enterprise text-to-SQL.
\Cref{sec:dataset} describes the dataset construction process, and \Cref{sec:benchmark} presents our evaluation setup and results. \Cref{sec:error} analyzes the remaining errors, and \Cref{sec:conclusion} concludes.
\section{Related work}
\label{sec:related}

\textbf{Text-to-SQL datasets.}
Early text-to-SQL research primarily focused on benchmarks such as Spider~\citep{yu2018spider}, WikiSQL~\citep{zhongSeq2SQL2017}, KaggleDBQA~\citep{kaggledbqa}, and MultiTabQA~\citep{pal2023multitabqa}, which are built on small, publicly available databases with relatively simple schemas and SQL queries of low to moderate complexity.
More recent benchmarks, including ScienceBenchmark~\citep{zhang2023sciencebenchmark}, BIRD~\citep{li2024can}, MMQA~\citep{wu2025mmqa}, and Spider 2.0~\citep{lei2025spider2}, aim to better reflect real-world scenarios by increasing database scale and query complexity. Despite these advances, existing datasets still rely on public databases and/ or queries derived from online tutorials, forums, or prior simplified benchmarks.
In contrast, our dataset is constructed from authentic query logs and tables drawn from private enterprise data warehouses. As a result, it captures the unique ``messiness'' of real-world production environments that are largely absent from existing benchmarks, including intricate schemas with non-obvious naming conventions, implicit domain knowledge, and production queries that extensively involve analytical constructs, advanced functions, deep nesting, and common table expressions.

\textbf{Text-to-SQL methods and evaluation.}
Traditional text-to-SQL research has largely evaluated methods using execution accuracy~\citep{yu2018spider, li2024can}. More recent work has recognized that realistic text-to-SQL involves multiple interdependent subtasks and has proposed specialized methods for individual subtasks, including multi-table retrieval~\citep{chen-etal-2024-table, chen2025can, wu2025mmqa}, schema linking~\citep{Li2023RESDSQLDS, Talaei2024CHESSCH}, and query decomposition~\citep{pourreza2023din, fan2024zeronl2sql}. These subtasks, however, are often evaluated in isolation or on benchmarks that do not fully capture enterprise-scale complexity.
Our work addresses this limitation by introducing a dataset and evaluation framework that supports holistic diagnosis.
\dataset{} introduces five critical subtasks: multi-table retrieval, join key detection, column mapping, domain knowledge extraction, and query decomposition.
Unlike existing benchmarks that provide only question-SQL pairs or partial annotations for a subset of subtasks, \dataset{} includes complete human annotations for all subtasks along with their evaluation metrics. This design enables fine-grained bottleneck analysis, clearly revealing where modern text-to-SQL frameworks~\citep{deng2025reforce, 10.14778/3641204.3641221} break down in complex enterprise environments.

\textbf{Query synthesis and generation.}
Prior work has explored the use of large language models (LLMs) for automated query synthesis to efficiently construct large-scale text-to-SQL datasets. Systems such as Sense~\citep{Yang2024SynthesizingTD}, OmniSQL~\citep{li2025omnisql}, SING-SQL~\citep{caferouglu2025sing}, and SQL-factory~\citep{li2025sql} directly generate SQL queries conditioned on a given schema, concrete values, and a specified complexity level (e.g., simple or complex). These approaches rely on the LLM's exposure to public databases and SQL patterns during training. Consequently, they are not well-suited for generating realistic queries over private enterprise schemas that are unseen by public LLMs.
Other approaches synthesize new queries by transforming existing queries. Text2SQL-Flow~\citep{Cai2025Text2SQLFlowAR} augments seed queries through operations such as modifying constants or adding predicates, while RingSQL~\citep{sterbentz2026ringsql} constructs a fixed set of templates and produces new queries solely by instantiating them with different concrete values, without composing or recombining templates to create deeper or more complex structures. Both methods are fundamentally constrained by the structural patterns present in the initial seeds or predefined templates.
Our approach differs by first extracting \emph{atomic} structural templates from real enterprise query logs and then \textit{iteratively composing} them. This enables systematic exploration of the combinatorial space of a schema, producing queries with substantially greater structural depth, such as multi-level nesting and CTEs, while maintaining fidelity by grounding every composition in patterns observed in real production workloads. Furthermore, by selectively combining templates, we can create queries that isolate individual challenges, for example, queries with domain knowledge but have low complexity. This enables fine-grained diagnostic evaluation, pursuing an objective distinct from prior work that primarily emphasizes efficiency or scalability.

\section{Subtasks of enterprise text-to-SQL}
\label{sec:subtasks}

As discussed in \Cref{sec:intro}, enterprise text-to-SQL poses a combination of challenges. Models must navigate large and heterogeneous data warehouses to retrieve a relevant subset of tables; infer join relationships in the absence of explicit foreign key constraints; map natural language concepts to obscure column names and domain-specific instances; and correctly reason over sophisticated queries that commonly involve deep nesting, common table expressions (CTEs), advanced functions, and other complex analytical constructs.
A model must simultaneously resolve all challenges to generate a correct query. However, the standard accuracy metric based on LLM-generated and gold SQL outputs is all-or-nothing. Therefore, evaluating and interpreting model performance along individual challenge dimensions is particularly difficult, since errors arising from different challenges are conflated in a single binary accuracy measure (1 if query outputs match, 0 if not). This coarse-grained evaluation also masks incremental progress, as improvements in one subtask (e.g., better table retrieval) may not be reflected in the final accuracy if other subtasks fail.

To enable fine-grained evaluation, we augment end-to-end text-to-SQL task with five \textit{subtasks} that are critical for successful query generation, as shown in Figure \ref{fig:example}.
While addressing the end-to-end task does not strictly require addressing these subtasks, they serve as valuable indicators of fine-grained performance and provide insights for potential improvements.
We first present the end-to-end task and then describe the subtasks.

\textbf{End-to-end text-to-SQL.}
Following the standard problem setup of text-to-SQL, the input to a LLM includes a natural language question and a database of tables, and the output is a SQL statement whose execution answers the question.
A database includes a set of tables. Each table includes a schema (that describes the names of and data types of each column) and instances of each table column.

\textbf{Multi-table retrieval.}
As discussed in Section~\ref{sec:intro}, enterprise databases typically contain a large number of tables. Moreover, enterprise user questions commonly require information from a substantially larger set of tables compared to those in existing datasets. Consequently, models need to identify the appropriate set of connected tables~\cite{chen-etal-2024-table,chen2025can} from the large database schema that is necessary for constructing an accurate SQL statement.
For example, to answer the user question shown in \Cref{fig:example}, the model need to retrieve five relevant tables from a large database. One of these tables is \texttt{FCLT\_ROOMS}, as the question explicitly asks to ``show all rooms in the Stata building.''
Formally, given a natural language question and a database schema, the model must identify a set of relevant tables sufficient to express the correct SQL query.

\textbf{Join key detection.}
In addition to identifying the relevant tables, models must determine how these tables are connected by identifying overlapping or semantically related columns that serve as join keys. In enterprise databases, such relationships are often implicit and not specified via foreign key constraints.
As illustrated in \Cref{fig:example}, connecting information between the tables \texttt{FCLT\_ROOMS} and \texttt{CATALOG\_SUBJECT\_OFFERED} involves recognizing that two significantly different columns \texttt{FCLT\_ROOMS.FCLT\_ROOM\_KEY} and \\
\texttt{CATALOG\_SUBJECT\_OFFERED.MEET\_PLACE}
are semantically relevant, exhibit overlapping instances, and thus constitute a valid join condition.
Formally, given a natural language question and a database, the model must identify the join keys that can establish connections among the retrieved tables.

\textbf{Column mapping.}
To generate a correct SQL query, models must also identify which table columns correspond to the information mentioned in the user question. This involves mapping natural language phrases to schema elements that may use opaque or abbreviated naming conventions.
In the example, answering the phrase ``\textit{all rooms in the Stata building}'' involves accessing the columns \texttt{FCLT\_ROOM\_KEY} and \texttt{FCLT\_BUILDING\_KEY} in the table \texttt{FCLT\_ROOMS}, which respectively represent room identifiers and building identifiers.
Formally, given a natural language question and a database, the model must map semantic units in the question (e.g., ``Stata building'') to the appropriate table columns (e.g., \texttt{FCLT\_ROOMS.FCLT\_BUILDING\_KEY}).

\textbf{Domain knowledge.}
Beyond identifying relevant columns, models must further map natural language entities to specific instance values stored in the database. This often involves leveraging implicit domain knowledge that is not directly available from a single table.
In the running example, identifying that ``Stata building'' corresponds to the column \texttt{FCLT\_ROOMS.FCLT\_BUILDING\_KEY} is insufficient on its own, as the table \texttt{FCLT\_ROOMS} does not store building names with their associated corresponding keys. The model must therefore explore other tables in the database and discover that the table \texttt{BUILDINGS} contains both columns \texttt{BUILDING\_KEY} and \texttt{BUILDING\_NAME}. It must further infer that “Stata building” semantically corresponds to \texttt{Ray and Maria Stata Center}, and thus determine that the appropriate building key is \texttt{32}, even though this table does not appear in the final SQL statement.
Formally, given a natural language question and a database, the model must identify the appropriate instance values, together with column mappings, that can be used to form predicates in the final SQL statement.

\textbf{Query decomposition.}
Finally, correctly translating enterprise user questions into SQL involves reasoning about question complexity and determining whether the query should be decomposed into multiple interdependent subproblems. Attempting to generate a complex SQL query in a single pass is often brittle and error-prone.
For the example question, the gold SQL exceeds 100 lines. A more reliable solution strategy is to decompose the question into sub-queries, including one computing department subject units and another identifying department rooms. These sub-queries can be solved independently and expressed as Common Table Expressions (CTEs), for example \texttt{department\_subject\_unit} and \texttt{department\_room}, which are then joined to produce the final result.
Formally, given a natural language question and a database, the model should identify a feasible solution plan. For simple questions, this may involve generating the SQL query directly, whereas for complex questions, it involves decomposing the problem into multiple sub-steps, solving them independently, and composing the results into a final SQL query.
\section{\dataset{} Dataset}
\label{sec:dataset}

As described in Section \ref{sec:intro}, existing text-to-SQL datasets are largely derived from public databases and online queries to reflect a relatively simplified setting that does not fully capture the high schema and query complexity found in private enterprise data warehouses.
To bridge this gap, we construct our dataset, \dataset{}, using three \textit{private enterprise} data warehouses.
We describe the data collection process in \Cref{sec:data-collection}, dataset expansion via Structural Template Recomposition in \Cref{sec:data-expansion}, and present overall dataset statistics in \Cref{sec:data-stats}.

\begin{figure}
\centering
\includegraphics[page=4, width=0.4\textwidth, trim={7cm 1cm 7cm 1cm}, clip]{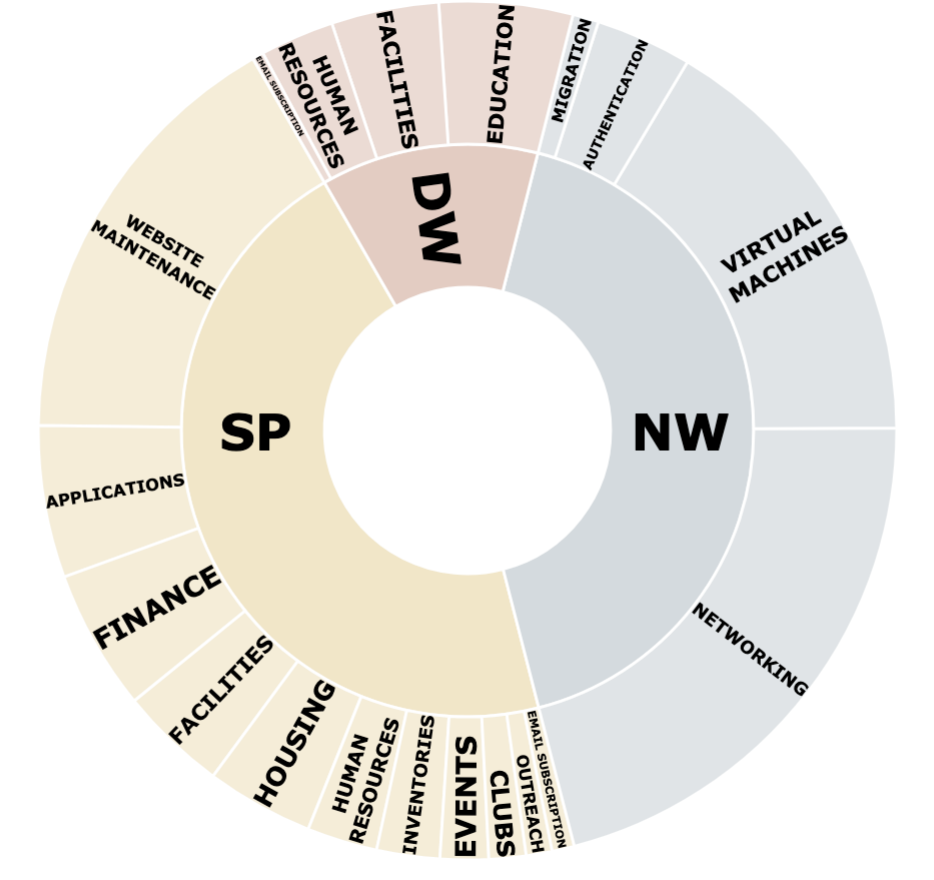}
\caption{\dataset{} contains tables from three private enterprise data warehouses, \texttt{DW}, \texttt{NW}, and \texttt{SP}, spanning 19 diverse domains, including facilities, education, and networking.}
\label{fig:domain}
\end{figure}






\begin{table*}[!ht]
\caption{\label{tab:example-annotation} The complete expert-verified annotation for the example shown in \Cref{fig:example}.}
\begin{tcolorbox}[]
\begin{lstlisting}[breaklines=true, basicstyle=\ttfamily\small]
{
  "Question": "For each of the top 10 departments in the HASS school ranked by total subject...",
  "SQL": "WITH department_subject_unit AS (SELECT DEPARTMENT_NAME, TOTAL_UNITS, ...",
  "Tables": ["SIS_SUBJECT_CODE", "SIS_COURSE_DESCRIPTION", "SUBJECT_OFFERED_SUMMARY", ...],
  "Join keys": [["FCLT_ROOMS.FCLT_ROOM_KEY", "CATALOG_SUBJECT_OFFERED.MEET_PLACE"], ...],
  "Column mapping": {
    "rooms": [FCLT_ROOMS.FCLT_ROOM_KEY], "Stata building": [FCLT_ROOMS.FCLT_BUILDING_KEY], ...
  },
  "Domain knowledge": ["Stata building refers to BUILDINGS.BUILDING_KEY = 32", ...],
  "Sub-questions": [
    "For each department, provide the department name, the range, variance, and standard deviation of total subject units, considering only subjects offered in the HASS school...",
    "For all rooms with an area greater than 400 in the Stata building, show the room key, building room, area, the subject of the course, ...", ...
  ],
  "Sub-SQLs": [
    "SELECT DEPARTMENT_NAME, MAX(TOTAL_UNITS) - MIN(TOTAL_UNITS), VARIANCE(TOTAL_UNITS), ...",
    "SELECT FCLT_ROOM_KEY, BUILDING_ROOM, AREA, SUBJECT_TITLE, ...", ...
  ]
}
\end{lstlisting}

\end{tcolorbox}
\end{table*}

\subsection{Dataset collection}
\label{sec:data-collection}

As discussed in \Cref{sec:subtasks}, the standard text-to-SQL task formulation involves databases and question-SQL pairs. To support fine-grained evaluation, we additionally introduce five subtasks critical to query generation. We describe the annotation process for each.

\textbf{Databases.}
We collected table information from three private data warehouses, including table names, column names, column types, rows, and join keys (if they exist). The first data warehouse, \texttt{DW}, consists of 97 tables and 1530 columns from an Oracle data warehouse operated by the IT department of an academic institution. It serves as the backbone for managing the institution's education systems, covering courses, students, faculty, and libraries, as well as facilities management, including buildings and rooms.
The second, \texttt{NW}, comprises 366 tables and 2708 columns spanning five MySQL databases. It is operated by the infrastructure team of a research laboratory and underpins the management of its compute cluster, including machines, virtualization, security groups, and identity and access management.
The third, \texttt{SP}, contains 349 tables and 2717 columns from two MySQL databases. It supports the housing management software of a residential facility, handling resident check-in and check-out, package tracking, and event management.
Collectively, these tables span 19 diverse domains with the distribution summarized in Figure \ref{fig:domain}.
In accordance with the privacy policies of the contributing organizations, all publicly released versions of the dataset undergo instance-level anonymization in both tables and queries to eliminate any personally identifiable information.

\textbf{Question-SQL pairs.}
To reflect the true complexity of queries posed on enterprise databases, we first collected real user query logs and reports from source organizations. We then extracted complete, semantically meaningful SQL statements that reflect real analytical use cases commonly found in these logs and reports.
The annotation of natural language questions for each SQL statement was conducted collaboratively by a team of six graduate students with formal training in database systems and six professional database administrators from the enterprise data warehouse support group. The administrators are domain experts who actively develop and use the databases and possess comprehensive knowledge of the underlying schemas, query logs, and business semantics.

Prior to annotation, all annotators reviewed representative examples from existing text-to-SQL datasets, including Spider~\citep{yu2018spider} and Bird~\citep{li2024can}, to establish consistent guidelines for writing natural language questions that balance naturalness with unambiguity. In addition, database administrators worked closely with the students to provide training on database schemas, table relationships, and domain-specific conventions.
For each SQL statement, annotators drafted corresponding natural language questions following strict guidelines: questions were required to eliminate ambiguity in expected results, explicitly reflect all predicates and information used in the SQL (while preserving their logical ordering), and maintain clarity. LLM-based checkers were used as auxiliary tools to help identify unclear or inconsistent phrasing, while final correctness and fidelity to the original SQL intent were ensured through human review. Each question was examined by at least two graduate students and two database administrators, and only those passing the majority voting were retained. This process resulted in an inter-annotator agreement of approximately 83\%.
Together, this expert-in-the-loop, consensus-driven process ensures that the resulting question-SQL pairs faithfully reflect real enterprise usage and capture the necessary domain-specific semantics.



\textbf{Subtasks.}
As described in \Cref{sec:subtasks}, we identify five subtasks critical to enterprise text-to-SQL. To enable fine-grained analysis of LLM performance beyond binary execution-accuracy measurements, we provide human annotations for each subtask.

Following the same rigorous protocol used for annotating question-SQL pairs, annotations for all subtasks were produced collaboratively by all annotators and subjected to the same multi-stage review process. Initial drafts were generated either manually or with the assistance of professional SQL parsers. Specifically, the tables and join keys involved in each SQL statement were extracted using established SQL parsing libraries, \href{https://github.com/tobymao/sqlglot}{\texttt{SQLGlot}} and \href{https://github.com/andialbrecht/sqlparse}{\texttt{sqlparse}}, and subsequently verified by human annotators.

Column mappings were manually annotated by pairing each topic phrase appearing in the user question with the corresponding table columns. Domain knowledge was annotated as a list of domain facts.
Query decomposition was annotated as a sequence of sub-questions and their associated sub-SQL statements, whose composition yields the final SQL query. After undergoing the same review and majority-voting process used for question-SQL annotation, these subtask annotations achieved an average inter-annotator agreement of approximately 87\%.
\Cref{tab:example-annotation} presents the fully expert-verified annotations for the example shown in \Cref{fig:example}. Due to security constraints, we were only able to obtain one month of query logs and reports from the source organizations. After manual review, we collected 254 fully-annotated queries.


\subsection{Dataset expansion}
\label{sec:data-expansion}

As discussed in \Cref{sec:data-collection}, although queries collected from real logs and reports yield high-quality question-SQL pairs, the volume of accessible data is inherently limited by the privacy and security constraints of participating organizations.
Moreover, while the subtasks introduced in \Cref{sec:subtasks} enable fine‑grained evaluation, we aim to complement this through dataset‑level augmentation.
Specifically, to address both data scarcity and the limitations of coarse‑grained evaluation, we expand the dataset using a systematic query synthesis pipeline termed \textit{Structural Template Recomposition}.

At a high level, the pipeline first extracts representative structural templates from real query logs. These templates are then instantiated with concrete values, such as tables, columns, and instances, and \textit{iteratively} composed to produce complex queries.
By selectively combining templates, we generate queries that isolate individual challenges: queries with domain knowledge but low complexity, queries with high complexity but no domain knowledge, and queries combining both domain knowledge and high complexity.
We focus on isolating domain knowledge and query complexity because these are query-level properties that can be systematically controlled during synthesis. In contrast, other challenges such as table retrieval, join key detection, and column mapping arise from schema-level characteristics, which we do not wish to modify in order to preserve the realism of the enterprise tables.
All synthesized queries are verified by the same group of expert annotators introduced in \Cref{sec:data-collection} to ensure correctness and fidelity to real enterprise query patterns. Overall, this pipeline both increases dataset scale and enables more fine-grained evaluation of model capabilities.
We next detail our pipeline by describing the extraction of templates from query logs and the iterative composition process used to generate new queries.

\begin{figure}[!htb]
\centering
\includegraphics[width=\linewidth, page=1, clip]{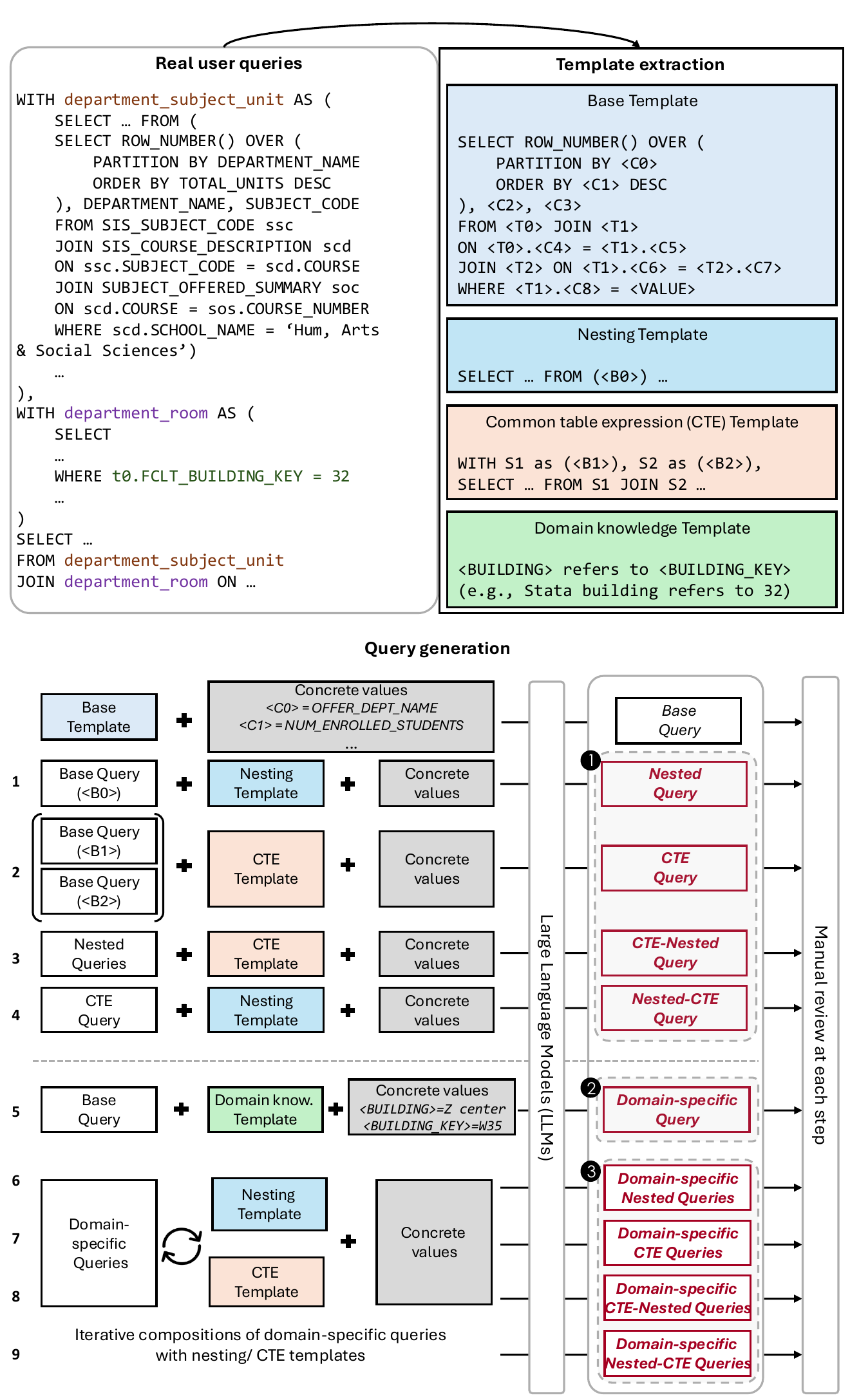}
\caption{Overview of the query synthesis pipeline: \textit{Structural Template Recomposition}. We first extract representative templates from real enterprise query logs, then instantiate them with concrete values. Templates are iteratively composed using LLMs to generate high-fidelity synthetic queries that focus on \protect\circled{1} query complexity, \protect\circled{2} domain knowledge, and \protect\circled{3} their combination. All generated queries are annotated and expert-verified using the process described in \Cref{sec:data-collection}.}
\label{fig:expansion}
\end{figure}

\textbf{Template extraction.}
By analyzing real enterprise query logs, we observe that most queries are composed of recurring analytical patterns centered around a
\texttt{SELECT-FROM-JOIN-WHERE-AGGREGATION} structure, augmented with varying degrees of nesting, Common Table Expressions (CTEs), and predicates grounded in domain-specific knowledge.
For example, the SQL statement shown in \Cref{fig:expansion} (introduced in \Cref{sec:intro}) contains a core analytical pattern combined with one level of nesting, two CTEs, and domain-specific predicates involving ``Stata building".
Motivated by this observation, we extract four types of templates from real queries:
\begin{itemize}
\item \textit{Base templates} ($T_B$), which capture the core\\
\texttt{SELECT-FROM-JOIN-WHERE-AGGREGATION} structure
\item \textit{Nesting templates} ($T_N$), which describe how subqueries are composed through nesting
\item \textit{CTE templates} ($T_C$), which capture how multiple subqueries are combined using CTEs
\item \textit{Domain knowledge templates} ($T_D$), which represent domain-specific predicates.
\end{itemize}
To obtain these templates, we scan each SQL statement in the query log and reports for snippets corresponding to base structures, nesting, CTE, or domain knowledge. All table names, column names, and literals are removed to produce abstract, reusable templates that can later be instantiated with different concrete values. We explicitly ensure that base, nesting, and CTE templates are free of domain knowledge. The extracted templates are then de-duplicated, and only those appearing frequently
are retained. Example templates are shown in \Cref{fig:expansion}.
Applying this process to real query logs yields a total of 594 highly representative templates.

\textbf{Query generation.}
Given the extracted templates, we synthesize queries in a bottom-up manner by instantiating them with concrete values and iteratively composing them using large language models (LLMs), with expert verification at each step, leading to queries that focus on domain knowledge, query complexity, and their combination.

\begin{table}[!htb]
    \centering
    \caption{Summary of Notation.}
    \label{tab:notations}
    \begin{adjustbox}{max width=\linewidth}
        \begin{tabular}{lp{8cm}}
                        \toprule
            Symbol & Description \\
            \midrule
            $\mathcal{M}$ & The LLM used for query synthesis. \\
            $\mathbf{v}$ & Concrete values sampled from the database (e.g., specific tables, columns, join keys). \\
            \midrule
            $T_B$ & Base templates capturing the core \texttt{SELECT-FROM-JOIN-WHERE-AGGREGATION} structure. \\
            $T_N$ & Nesting templates describing how subqueries are composed through nesting. \\
            $T_C$ & Common Table Expression (CTE) templates capturing how multiple subqueries are combined using CTEs. \\
            $T_D$ & Domain knowledge templates encoding domain-specific predicates. \\
            \midrule
            $Q_B$ & Base queries generated by instantiating $T_B$; used as intermediate building blocks. \\
            $Q_N$ & Nested queries generated by composing $T_N$ with $Q_B$. \\
            $Q_C$ & CTE queries generated by composing $T_C$ with multiple $Q_B$. \\
            $Q_{NC}$ & Nested-CTE queries generated by composing $T_N$ with $Q_C$. \\
            $Q_{CN}$ & CTE-nested queries obtained by composing $T_C$ with multiple $Q_N$. \\
            \midrule
            $Q_X^D$ & Domain-specific variant of $Q_X$ (domain knowledge is introduced via instantiating $T_D$), for $X\in\{B,N,C,NC,CN\}$. \\
            \bottomrule
        \end{tabular}
    \end{adjustbox}
\end{table}

\begin{table*}[!htb]
\centering
\caption{Database and query statistics of \dataset{} and other text-to-SQL datasets, along with the number of annotated subtasks (out of the five subtasks defined in \Cref{sec:subtasks}). *denotes statistics computed from the subset of gold SQLs that are available, since the full set is hidden.
\textbf{Bold} denotes the highest values.
}
\begin{adjustbox}{max width=\linewidth}
\begin{tabular}{c|ccc|ccccccc|c}
Dataset
& \makecell{\#Table\\/DB} & \makecell{\#Cols\\/DB} & \makecell{Join degree\\/DB}
& \makecell{\#Test\\queries}
& \makecell{\#Tokens\\/query}
& \makecell{\#Tables\\/query}
& \makecell{\#Joins\\/query}
& \makecell{\#Functions\\/query}
& \makecell{\#Nesting\\/query}
& \makecell{\#CTE\\/query}
& \makecell{\#Annotated\\subtasks}
\\
\midrule
WikiSQL & 1.0 & 6.3 & 0.0 & 15878 & 10.4 & 1.0 & 0.0 & 0.3 & 1.0 & 0.0
& 0/5\\
Spider & 5.1 & 25.7 & 1.6 & 2147 & 22.0 & 1.6 & 0.6 & 0.6 & 1.1 & 0.0 & 0/5\\
BIRD & 6.8 & 72.5 & 2.5 & 1534 & 37.3 & 2.0 & 0.9 & 1.3 & 1.1 & 0.0 & 1/5\\
MMQA & 2.3 & 12.4 & 1.1 & 3313 & 35.3 & 2.2 & 1.1 & 0.5 & 1.2 & 0.0 & 2/5\\
Spider 2.0 & 52.6 & 803.6 & 0.9* & 547 & 144.5 & 2.9* & 2.4* & 6.5 & 4.9* & 2.5* & 1/5\\
\midrule
\dataset{} & \textbf{101.5} & \textbf{869.4} & \textbf{9.8} & 9128 & \textbf{316.7} & \textbf{4.0} & \textbf{5.7} & \textbf{8.0} & \textbf{5.6} & \textbf{3.7} & \textbf{5/5}\\
\bottomrule

\end{tabular}
\end{adjustbox}

\label{tab:stats}
\end{table*}

We use the advanced model \texttt{GPT-5.2} to generate initial drafts of synthetic queries. The model, denoted as $\mathcal{M}$, is prompted with (1) a template, (2) a set of concrete values $\mathbf{v}$, and, when applicable, (3) previously synthesized queries.
The values $\mathbf{v}$ include specific tables, columns, join keys, instances, and literals. They are created by randomly sampling from the full database schema. Invalid table combinations without join keys are filtered out.
During the iterative composition stage, $\mathbf{v}$ is generated with awareness of previously synthesized queries, ensuring that tables and other schema elements used in earlier queries are consistently inherited rather than replaced with unrelated ones.
This setup ensures that generated queries are structurally valid and grounded in real schema elements.

Using the notation shown in Table \ref{tab:notations}, base queries $Q_B$ are obtained by instantiating base templates $T_B$ with concrete values $\mathbf{v}_1$: $$\mathcal{M}(T_B, \mathbf{v}_1) \to Q_B$$
where $\mathbf{v}_1$ may include, for example, \texttt{<C0> = OFFER\_DEPT\_NAME} and \texttt{<C1> = NUM\_ENROLLED\_STUDENTS}, as illustrated in \Cref{fig:expansion}.
Before being used as inputs for further composition, all base queries are manually reviewed following the protocol described in \Cref{sec:data-collection}. This review includes correctness and clarity checks, as well as execution to ensure non-empty results, and serves to prevent error propagation in subsequent steps.

Using the expert-verified base queries, we \textit{iteratively} synthesize more complex queries by composing them with nesting templates $T_N$ and CTE templates $T_C$:
$$\mathcal{M}(T_N, Q_B, \mathbf{v}_2) \to Q_N,\quad \mathcal{M}(T_C, \{Q_B\}, \mathbf{v}_3) \to Q_C$$
where $T_C$ takes multiple base queries as input, and $Q_N$ and $Q_C$ denote nested and CTE queries, respectively. After expert verification, these queries can be further composed:
$$\mathcal{M}(T_N, Q_C, \mathbf{v}_4) \to Q_{NC},\quad \mathcal{M}(T_C, \{Q_N\}, \mathbf{v}_5) \to Q_{CN}$$
where $Q_{NC}$ and $Q_{CN}$ denote nested-CTE queries and CTE-nested queries, respectively.
All composed queries are again reviewed using the same verification procedure.
To avoid generating excessively complex or unnatural queries, we restrict synthesis to at most two levels of nesting or CTE chaining, as deeper compositions are empirically difficult for models to synthesize correctly.

In parallel, we incorporate domain knowledge into queries by instantiating domain knowledge templates $T_D$ with concrete domain facts and attaching them as predicates to base queries. Given a base query $Q_B$, we obtain a domain-specific variant $Q_B^D$ as
$$\mathcal{M}(T_D, Q_B, \mathbf{v}_6) \to Q_B^D$$
where $\mathbf{v}_6$ may include domain facts such as \texttt{Z center refers to building key W35}. Domain-specific queries can then undergo the same composition process outlined above:
$$\mathcal{M}(T_N, Q_B^D, \mathbf{v}_7) \to Q_N^D \quad \mathcal{M}(T_C, \{Q_{B}^D\}, \mathbf{v}_8) \to Q_C^D$$
and further to $Q_{NC}^D$ and $Q_{CN}^D$, where $Q_X^D$ denotes $Q_X$ with domain knowledge.

This pipeline substantially increases the dataset scale. Importantly, because all templates are extracted from real enterprise query logs, the synthesized queries preserve the structural and semantic characteristics of real-world workloads.
Moreover, the bottom-up construction with selective combination of templates naturally yields three categories of queries: \protect\circled{1} complex queries without domain knowledge ($Q_N, Q_C, Q_{NC}, Q_{CN}$), \protect\circled{2} domain-specific queries with low complexity ($Q_B^D$), and \protect\circled{3} queries that jointly stress high complexity and domain knowledge ($Q_N^D, Q_C^D, Q_{NC}^D, Q_{CN}^D$). This enables targeted evaluation and facilitates the isolation of individual challenges faced by text-to-SQL methods in enterprise settings.

Base queries $Q_B$ are used only as intermediate building blocks and are not included in the final dataset. Overall, the pipeline produces 8874 high-fidelity, expert-verified SQL queries. Annotations for each SQL query, including natural language questions and subtasks, are generated using the same procedure described in \Cref{sec:data-collection}.


\subsection{Dataset statistics}
\label{sec:data-stats}

Table~\ref{tab:stats} compares the statistics of \dataset{} databases, queries, and subtask annotations with widely used text-to-SQL benchmarks, including WikiSQL~\citep{zhongSeq2SQL2017}, Spider~\cite{yu2018spider}, BIRD~\cite{li2024can}, as well as more recent datasets including MMQA~\citep{wu2025mmqa} and Spider~2.0~\citep{lei2025spider2}. The comparison highlights the substantially higher complexity of both the underlying databases and the query workloads in \dataset{}, reflecting the challenges inherent to real-world private enterprise data warehouses. It also shows that \dataset{} provides a richer set of subtask annotations, enabling more fine-grained analysis.

\textbf{Database statistics.}
We observe that the databases in \dataset{} are markedly more complex than those in existing benchmarks, both in terms of the number of tables and the number of columns per database. As discussed in \Cref{sec:intro}, this significantly increases the difficulty of text-to-SQL, as models must accurately retrieve a relevant subset of tables and identify the correct columns from a much larger and denser schema space.
Beyond schema size, we also report the average \emph{join degree} across all tables, where the join degree of a table is defined as the number of other tables it can be joined with. A higher join degree implies a greater number of possible join paths, substantially increasing the risk of selecting semantically incorrect joins. This characteristic is especially pronounced in \dataset{}, where tables are often highly interconnected and encode rich domain semantics that are not explicitly captured by schema metadata alone.

\textbf{Query statistics.}
In addition to database complexity, the queries in \dataset{}, derived from real-world enterprise query logs, are significantly more challenging than those in prior benchmarks. As shown in \Cref{tab:stats}, these queries are significantly longer, with much higher average token counts. They typically involve retrieving and joining a larger number of tables to satisfy complex information needs. Furthermore, these queries exhibit more frequent use of advanced functions (window, aggregate, scalar), deeper nesting, and extensive use of Common Table Expressions (CTEs). Together, these characteristics reflect the structural sophistication of real enterprise SQL and underscore the gap between existing text-to-SQL benchmarks and practical deployment settings.
Moreover, \dataset{} includes a large number of complex queries, exceeding the scale of recent complex benchmarks such as Spider~2.0. This breadth provides a more comprehensive testbed for evaluating and developing models capable of handling diverse enterprise query patterns.

\textbf{Subtask annotations.}
As detailed in \Cref{sec:subtasks,sec:data-collection}, we introduce five subtasks and provide full annotations for each. In contrast to prior datasets, which typically annotate only question–SQL pairs and a limited subset of subtasks, \dataset{} enables more comprehensive fine-grained error analysis, allowing clearer identification of the root causes of errors.

\section{Evaluation}
\label{sec:benchmark}

As discussed in \Cref{sec:dataset}, in addition to standard question-SQL pairs, \dataset{} provides subtask annotations and three query categories designed to isolate query complexity and domain knowledge challenges. With these in place, we outline the evaluation process.

We first assess how existing text-to-SQL methods perform on \dataset{} using standard, coarse-grained evaluation metrics. However, as noted in \Cref{sec:intro}, such metrics alone make it difficult to interpret method behavior. To gain deeper insight into where current methods struggle, we therefore conduct fine-grained analyses based on subtask annotations and query categories. We outline the research questions (RQ) below:
\textbf{RQ1}: What is the end-to-end performance on \dataset{} using state-of-the-art LLMs and text-to-SQL methods?
\textbf{RQ2}: How do methods perform across query categories that emphasize different types of challenges?
\textbf{RQ3}: What is the fine-grained performance on each subtask?
\textbf{RQ4}: How does performance change when text-to-SQL methods are augmented with subtask annotations as oracle hints?

We begin by outlining the evaluation setup and metrics in \Cref{sec:exp-setup} and \Cref{sec:exp-metrics}.
We then present end-to-end performance without subtask annotations (RQ1-3) in \Cref{sec:exp-ete}, followed by results with subtask annotations (RQ4) in \Cref{sec:exp-hint}.

\subsection{Evaluation setup}
\label{sec:exp-setup}

We evaluate text-to-SQL methods under two settings: (1) standard end-to-end SQL generation without subtask annotations, and (2) SQL generation augmented with subtask annotations as hints.

\textbf{End-to-end SQL generation.}
Following leaderboards from recent text-to-SQL benchmarks such as BIRD~\cite{li2024can} and Spider~2.0~\cite{lei2025spider2}, we select a set of state-of-the-art (SOTA) LLMs and text-to-SQL methods.
LLMs include both proprietary and open-source models:
\begin{itemize}
\item Proprietary models: \texttt{GPT-\{o4-mini, 5.2, 5-mini\}} (with \texttt{GPT-o4-mini} serving as an advanced reasoning model)~\cite{singh2025openai}, \texttt{Claude-4.5-sonnet}, and \texttt{Gemini-2.0-flash}.
\item Open-source models: \texttt{Qwen3-Next-80B-A3B-Instruct}~\cite{qwen3technicalreport} and \texttt{MiniMax M2.1}.
\end{itemize}
Text-to-SQL methods include \reforce{}~\citep{deng2025reforce}, DIN-SQL~\citep{pourreza2023din}, and DAIL-SQL~\citep{10.14778/3641204.3641221}.
\reforce{} is an agentic method leveraging candidate generation, majority voting, and column exploration.
DIN-SQL uses query decomposition and self-correction.
DAIL-SQL applies prompt engineering and example selection.
We also evaluate few-shot prompting, where models receive a small set of demonstrations before generating SQL.
As shown in \Cref{tab:stats}, \dataset{} databases contain many tables and columns, making it infeasible to include full schemas within model context windows. To address this, we use retrieval-augmented generation~\citep{lewis2020retrieval}, providing each method with the top‑$k$ tables that are most semantically similar to the user question. We retrieve these tables using SOTA dense embeddings (\texttt{Qwen3-Embedding-8B}~\cite{qwen3embedding}) and then rerank them using a strong reranker (\texttt{Qwen3-Reranker-8B}).

\textbf{SQL generation with subtask annotation.}
In this setting, we provide methods with subtask annotations defined in \Cref{sec:subtasks} to evaluate how performance changes when methods are augmented with oracle hints.
This setup mirrors the end-to-end setting, except that any provided subtask annotation is inserted directly into the model prompt/ context to assist models with solving the subtask.
For example, when providing table retrieval annotations, instead of giving the method top-$k$ retrieved tables, we provide the exact set of tables used in the gold SQL statement.

\begin{table*}
\caption{End-to-end execution accuracy on \dataset{}.}
\centering
\begin{adjustbox}{max width=\linewidth}
\begin{tabular}{lcccccccc}
& \multicolumn{5}{c}{Proprietary models} & \multicolumn{2}{c}{Open-source models} \\
\cmidrule(lr){2-6} \cmidrule(lr){7-8}
& \makecell{Claude-4.5-\\sonnet} & \makecell{GPT-o4-mini\\(Reasoning model)} & GPT-5.2 & GPT-5-mini & \makecell{Gemini-2.0-\\flash} & \makecell{Qwen3-Next-80B-\\A3B-Instruct} & MiniMax M2.1 & Overall \\
\midrule
\reforce{} & 11.4 & 10.6 & 10.8 & 8.4 & 8.5 & 9.0 & 7.7 & 9.5 \\
Few-shot   & 8.8  & 8.7  & 7.6  & 7.9 & 6.7 & 6.7 & 7.0 & 7.6 \\
DIN-SQL    & 5.7  & 6.1  & 6.1  & 5.6 & 3.7 & 4.2 & 4.2 & 5.1 \\
DAIL-SQL   & 4.5  & 4.3  & 4.7  & 3.9 & 2.5 & 2.7 & 2.3 & 3.6 \\
\bottomrule
\end{tabular}
\end{adjustbox}
\label{tab:acc}
\end{table*}

\subsection{Evaluation metrics}
\label{sec:exp-metrics}

We outline the coarse-grained execution accuracy metric commonly used in existing text-to-SQL benchmarks, followed by a description of our fine-grained metrics for the annotated subtasks.

\textbf{Execution accuracy.}
Prior text-to-SQL benchmarks~\cite{yu2018spider, li2024can} primarily assess end-to-end performance using execution accuracy. This metric assigns a score of 1 when the method-generated SQL returns the same execution result as the gold SQL, and 0 otherwise. As discussed in~\Cref{sec:intro}, this binary, all-or-nothing metric is coarse-grained and offers limited interpretability, making it difficult to diagnose sources of method errors. This motivates the fine-grained evaluation based on subtasks described in~\Cref{sec:subtasks}.

\textbf{Table retrieval F1.}
To measure a method's ability to identify the correct set of tables, we compute the F1 score between the set of tables referenced in the method-generated SQL statement and those in the gold SQL. Table sets are extracted using professional SQL parsers (\texttt{SQLGlot} and \texttt{sqlparse}) and LLMs.

\textbf{Column mapping F1.}
We further evaluate whether the method selects the correct set of columns. This metric uses an F1 score computed over the set of columns appearing in the LLM-generated SQL versus those in the gold SQL. Columns are extracted as fully qualified identifiers of the form \texttt{table.column} using parsers and LLMs.

\textbf{Join key F1.}
This metric assesses the method's ability to identify correct join keys between tables. We compute the F1 score over the set of join conditions found in the LLM-generated SQL and those in the gold SQL. Join keys are extracted as expressions of the form \texttt{table1.column1 = table2.column2} using parsers and LLMs. Each join is treated as an unordered pair to ensure the metric is invariant to ordering.

\textbf{Domain knowledge F1.}
We evaluate a method's ability to produce domain-specific predicates. We compute the F1 score over the predicate sets extracted from the LLM-generated and gold SQL statements. Predicates are extracted as expressions of the form \texttt{table.column <OP> value} using parsers and LLMs.

\textbf{Query decomposition score.}
Query decomposition assesses a method’s ability to break a complex natural language question into meaningful intermediate sub-SQL statements. Because this subtask allows for substantial variation in valid decomposition paths, we adopt an LLM-as-a-Judge~\cite{zheng2023judging} approach as the primary evaluation metric.
Specifically, we provide an LLM (GPT‑5-mini in our case) with the LLM-generated SQL and the human-annotated sub‑SQLs. The judge model is prompted to assign a 0-5 score~\cite{Li2026GradingSI} based on a human-designed rubric that emphasizes how closely the LLM-generated SQL aligns \textit{semantically} with the human-annotated sub‑SQLs. To align with the scales used for other evaluation metrics, accuracy and F1, we present the score normalized to the 0-100 interval.

\begin{table*}
\caption{Average end-to-end execution accuracy for each query category across all models.}
\centering
\begin{adjustbox}{max width=\linewidth}
\begin{tabular}{lcccc|c}
& \multicolumn{4}{c}{Synthetic queries} & Real queries\\
\cmidrule(lr){2-5} \cmidrule(lr){6-6}
& \makecell{\protect\circled{1} Complex queries
}
& \makecell{\protect\circled{2} Domain-specific queries
}
& \makecell{\protect\circled{3} Domain-specific complex queries
}
& Overall
& Overall
\\
\midrule
\reforce{}  & 7.2 & 32.0 & 5.4 & 9.5 & 8.2\\
Few-shot & 5.0 & 30.8 & 3.7 & 7.6 & 5.8\\
DIN-SQL  & 2.9 & 21.4 & 2.8 & 5.1 & 3.3\\
DAIL-SQL & 1.7 & 15.3 & 2.1  & 3.6 & 3.1\\
\bottomrule
\end{tabular}
\end{adjustbox}
\label{tab:acc-category}
\end{table*}

\begin{table*}
\caption{Avg. execution accuracy and subtask performance in end‑to‑end setting and with subtask annotations across all models.}
\centering
\begin{adjustbox}{max width=\linewidth}
\begin{tabular}{lcccccc}
& Execution Accuracy & Table retrieval F1 & Join key F1 & Column mapping F1 & Domain knowledge F1 & Query decomposition score\\
\midrule
\multicolumn{6}{l}{\textit{Setting 0: End-to-end (no subtask annotations)}}\\
\midrule
\reforce{} & 9.5 & 70.1 & 35.5 & 61.6 & 20.4 & 50.8 \\
Few-shot & 7.6 & 66.2 & 33.9 & 57.3 & 16.3 & 52.1 \\
DIN-SQL & 5.1 & 66.0 & 33.3 & 55.2 & 16.6 & 53.5 \\
DAIL-SQL & 3.6 & 59.3 & 31.2 & 51.2 & 16.7 & 49.9 \\
\midrule
\multicolumn{6}{l}{\textit{Setting 1: With annotations of schema-linking subtasks: Table retrieval, Join keys, Column mapping}}\\
\midrule
\reforce{} & 18.9 & 91.3 & 71.9 & 82.1 & 23.9 & 58.3 \\
Few-shot & 17.0 & 89.8 & 77.5 & 81.4 & 23.7 & 62.8 \\
DIN-SQL & 9.3 & 87.9 & 75.0 & 78.4 & 19.8 & 56.4 \\
DAIL-SQL & 8.0 & 88.6 & 70.9 & 77.3 & 23.1 & 57.3 \\
\midrule
\multicolumn{6}{l}{\textit{Setting 2: With annotations of all subtasks}}\\
\midrule
\reforce{} & 25.9 & 91.9 & 75.7 & 83.4 & 27.1 & 64.8 \\
Few-shot & 23.1 & 90.5 & 80.3 & 82.4 & 26.1 & 66.6 \\
DIN-SQL & 13.9 & 89.4 & 77.9 & 80.4 & 26.1 & 62.3 \\
DAIL-SQL & 11.2 & 89.3 & 74.7 & 79.4 & 26.4 & 63.3 \\
\bottomrule
\end{tabular}
\end{adjustbox}
\label{tab:acc-hint}
\end{table*}

\subsection{End-to-end performance}
\label{sec:exp-ete}

We investigate the performance on \dataset{} under the end-to-end setting from three perspectives:
(RQ1) execution accuracy on SOTA LLMs and text-to-SQL methods
(RQ2) execution accuracy on different query categories that isolate individual challenges, and
(RQ3) fine-grained evaluation based on subtask annotations.

\textbf{RQ1: Execution accuracy.}
\Cref{tab:acc} reports execution accuracy under the end‑to‑end setting. We observe that the state-of-the-art agentic text-to-SQL method, \reforce{} with Claude‑4.5‑Sonnet, achieves only 11.4\% accuracy on \dataset{}. In contrast, the same method attains 62.9\% accuracy~\cite{deng2025reforce} on Spider 2.0 (already considered a substantially more challenging benchmark than Spider and BIRD). This substantial drop highlights the difficulty and distinct characteristics of private enterprise text-to-SQL.
We also find that larger models (e.g., \texttt{Claude‑4.5‑Sonnet}, \texttt{GPT‑o4‑mini}) outperform smaller models (e.g., \texttt{GPT‑5‑mini}, \texttt{Gemini‑2.0‑Flash}), and proprietary models generally surpass open-source ones.
Across baselines, \reforce{} consistently delivers the strongest performance, followed by few‑shot prompting, then DIN‑SQL and DAIL‑SQL.
DIN‑SQL and DAIL‑SQL perform the worst despite employing advanced techniques such as query decomposition and example selection. This is largely due to their reliance on simplified, hard‑coded examples designed for benchmarks like Spider and BIRD. Such examples do not transfer well to the complex queries in \dataset{}, limiting the methods’ ability to adapt. In contrast, \reforce{} leverages agentic reasoning to explore the schema dynamically, making it more robust and generalizable across varying levels of difficulty. These results underscore the need for agentic methods capable of grounding reasoning directly in the schema to achieve broad generalization.


\textbf{RQ2: Execution accuracy of different query categories.}
Execution accuracy provides a useful high‑level metric, but it does not expose the root causes of errors. As outlined in \Cref{sec:data-expansion}, \dataset{} introduces three query categories designed to isolate specific subtask challenges: \protect\circled{1} complex queries without domain knowledge, \protect\circled{2} domain‑specific queries with minimal complexity, and \protect\circled{3} complex queries with domain knowledge.
\Cref{tab:acc-category} reports execution accuracy for each category, demonstrating the strengths and weaknesses of existing methods and pointing to the challenges that must be addressed. The results show that:
(1) all methods struggle substantially on both \protect\circled{1} and \protect\circled{2}. However, their significantly poorer performance on \protect\circled{1} indicates that constructing complex SQL structures is more challenging than resolving domain knowledge alone; and
(2) \protect\circled{3} is the most difficult overall, as it combines both structural complexity and domain‑specific reasoning.
\Cref{tab:acc-category} also includes results for real queries collected directly from the query logs described in \Cref{sec:data-collection}. The execution accuracy on real and synthetic queries is comparable, and the relative ranking of methods remains consistent. \textit{This provides further evidence that our synthesis pipeline (described in \Cref{sec:data-expansion}) produces high‑fidelity tasks that closely reflect the distribution of real‑world queries.}

\textbf{RQ3: Subtask performance.}
Moreover, we analyze the fine‑grained performance of existing methods across subtasks to better contextualize the low execution accuracy and identify areas that need improvement. \Cref{tab:acc-hint} reports both the end‑to‑end (\textit{Setting 0}) execution accuracy and subtask performance. Overall, subtask performance is far from satisfactory. In particular, methods perform worst on join key detection, domain knowledge, and query decomposition.
Join key detection is especially challenging because enterprise schemas often lack explicit join relationships, forcing methods to infer them during generation. Cryptic table and column names further obscure which attributes are joinable. The poor performance on domain knowledge and query decomposition similarly reflects current methods’ limited ability to infer domain conventions and break complex queries into coherent substeps. Together, these shortcomings highlight key characteristics of enterprise text‑to‑SQL, outlined in \Cref{sec:intro}, that require targeted advancements.

\subsection{Performance with subtask annotations}
\label{sec:exp-hint}

\Cref{sec:exp-ete} shows the low end-to-end performance, underscoring the difficulty of enterprise text-to-SQL. To probe the limits of current methods, we investigate \textbf{RQ4}: performance with subtask annotations as oracle hints. This setup serves two purposes: (1) it quantifies the attainable performance when methods are assisted in solving the subtask challenges, and (2) it identifies the residual errors not explained by those subtasks. Together, these results clarify both how much headroom is attributable to subtask weaknesses and what failure modes remain beyond them.

We evaluate two settings for providing subtask annotations.
\textit{Setting 1} provides annotations for schema-linking subtasks: table retrieval, join key detection, and column mappings.
\textit{Setting 2} provides annotations for all subtasks.
This progressive disclosure allows us to quantify how performance changes with increasing levels of guidance. Results are presented in \Cref{tab:acc-hint}.

We highlight three main observations:
(1) Providing annotations for schema‑linking subtasks leads to substantial gains in both execution accuracy and subtask performance. Providing annotations for all subtasks yields an additional boost, indicating that if methods could autonomously resolve these subtasks, end‑to‑end performance would improve significantly.
(2) Under \textit{Setting 2}, schema‑linking subtasks achieve over 75\% F1, yet domain knowledge F1 and query decomposition score remain low. This shows that the hardest challenges in \dataset{} arise from domain‑specific reasoning and the high structural complexity of enterprise queries.
(3) Even with annotations for every subtask, execution accuracy and several subtask metrics fall short of perfection. This suggests additional failure modes, which we analyze further in \Cref{sec:error}.
\section{Error Analysis}
\label{sec:error}


\begin{figure}
\centering
\includegraphics[page=1, width=\linewidth, trim=2.5cm 2.5cm 2.5cm 2.5cm, clip]{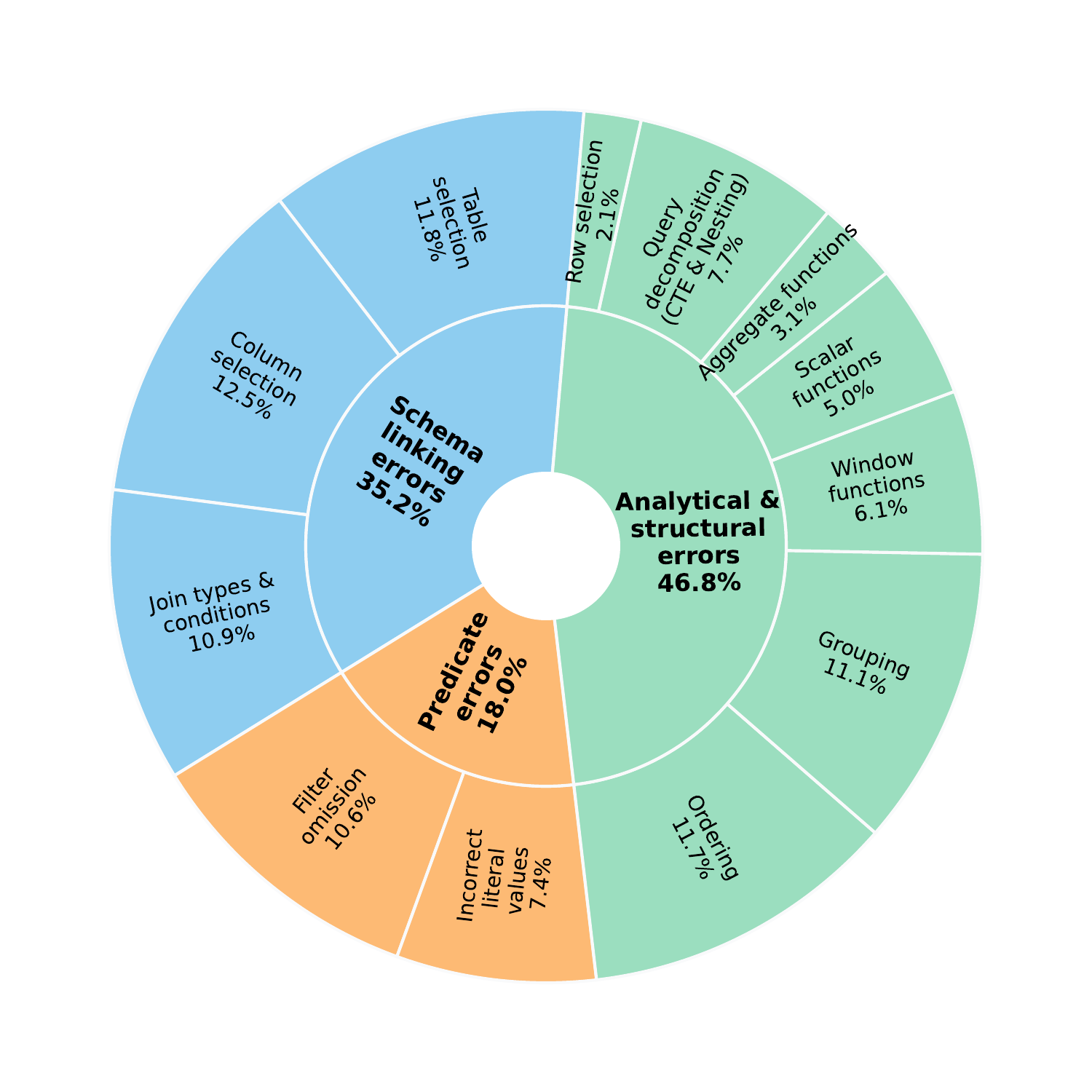}
\caption{Error distribution of text‑to‑SQL methods when subtask annotations are provided.}
\label{fig:error-analysis}
\end{figure}

In \Cref{sec:benchmark}, we evaluate both performance with and without subtask annotations. Although subtask annotations provide substantial improvements, overall performance could still be improved, as shown in \Cref{tab:acc-hint}.
To better understand the remaining challenges, we randomly sample 275 incorrect queries generated by text-to-SQL methods and analyze the residual errors.
The error categories and their distributions are summarized in \Cref{fig:error-analysis}. At a high level, three core issues emerge: mistakes in assembling the necessary logical structure required for a correct SQL query beyond query decomposition (analytic and structural errors), difficulties in correctly identifying schema elements (schema linking errors), and failures to capture or apply the appropriate filters (predicate errors), even when subtask annotations are provided. 

\textbf{Analytical and structural errors (46.8\%).}
As introduced in \Cref{sec:intro} and reflected in \Cref{tab:stats} statistics, \dataset{} includes highly complex enterprise‑grade analytical queries. They often involve advanced SQL functions and analytical constructs. These complexities give rise to several classes of errors beyond query decomposition:

(1) Grouping and ordering errors (22.8\%). Methods frequently omit or mis-specify \texttt{GROUP BY} or \texttt{ORDER BY} clauses, or place the \texttt{HAVING} clause incorrectly. Such mistakes lead to aggregations computed at the wrong granularity or filters applied at the wrong stage (e.g., applying a filter in the \texttt{WHERE} clause that should be applied after aggregation).

(2) Errors involving advanced functions (window, aggregate, and scalar) (14.2\%).
Methods often choose incorrect window functions, such as wrong \texttt{PARTITION BY}, \texttt{ORDER BY}, or \texttt{ROWS}/ \texttt{RANGE BETWEEN}, which distort row-level computations like cumulative aggregates or ranking operations. They may also select incorrect aggregate functions (e.g., using \texttt{COUNT} instead of \texttt{SUM} to compute total enrollment) or mix inconsistent variance/ standard‑deviation functions.
For scalar functions, methods may misuse or partially omit necessary expressions. For example, when a geometric mean should be computed using \texttt{EXP(AVG(LN(...)))}, methods may drop the \texttt{EXP}/ \texttt{LN} functions. Methods also occasionally insert unnecessary \texttt{CAST} operations, causing rounding artifacts or numeric drift.

(3) Query decomposition errors (7.7\%).
Even with subtask annotations for query decomposition, methods can still mis-handle advanced constructs. A common issue is incorrect or missing use of \texttt{ROLLUP}: for example, when the gold SQL uses \texttt{ROLLUP} to produce subtotals and a grand total row, methods may instead use a flat \texttt{GROUP BY}, thereby omitting required summary rows.

(4) Row selection errors (2.1\%).
Methods sometimes add or omit \texttt{DISTINCT}, apply incorrect \texttt{LIMIT} clauses, or introduce redundant deduplication steps. These mistakes alter the intended cardinality of the result set: either inflating it by failing to deduplicate or collapsing it by over-aggregating or over-restricting rows.

\textbf{Schema linking errors (35.2\%).}
Methods also frequently fail to select the correct tables, columns, and join conditions. Table‑selection and column‑selection errors account for 11.8\% and 12.5\% of cases, respectively. These challenges largely stem from the substantial number of tables and columns present in each database (\Cref{tab:stats}) and from opaque, domain‑specific naming conventions, for example, \texttt{SIS\_SUBJECT\_CODE} or \texttt{FCLT\_ROOMS}, which provide limited semantic guidance.
Additionally, methods often mis-infer join conditions or choose incorrect join types. For instance, a query may require a \texttt{LEFT JOIN} to preserve unmatched records, but methods may produce an \texttt{INNER JOIN}, thereby discarding essential rows. Such join‑related errors contribute another 10.9\%.

\textbf{Predicate errors (18.0\%).}
Methods exhibit two major forms of predicate‑level mistakes: generating predicates with incorrect literal values (7.4\%) and omitting required predicates altogether (10.6\%). These errors point to failures in domain‑specific reasoning. For example, a question requiring filters for privately owned academic buildings (\texttt{BUILDING\_TYPE = `ACADEMIC'} and \texttt{OWNERSHIP\_TYPE = `OWNED'}) may be answered by the method with SQL that excludes both conditions, returning results for all buildings. This often reflects the method’s limited familiarity with enterprise‑specific conventions and its inability to reliably include them even when subtask annotations are provided.

\section{Conclusion}
\label{sec:conclusion}




We present \dataset{}, the first text‑to‑SQL benchmark constructed from private enterprise data warehouses, addressing a critical gap left by existing benchmarks built on clean, public databases with relatively simple queries. By drawing from real‑world query logs, \dataset{} captures the true complexity of enterprise environments, including intricate schemas, domain‑specific conventions, and analytical constructs involving deep nesting, CTEs, and advanced functions.
To overcome the inherent scarcity of enterprise query logs, we augment the dataset with high‑fidelity, expert‑validated synthetic queries that isolate or combine core challenges such as domain knowledge and query complexity. Complementing this, we introduce a fine‑grained evaluation framework with human annotations for five critical subtasks, enabling diagnosis of model behavior beyond traditional execution accuracy.
Our comprehensive evaluation reveals that state‑of‑the‑art text‑to‑SQL systems perform dramatically worse on \dataset{} than on existing benchmarks, with advanced agentic frameworks using \texttt{GPT‑5.2} reaching only 10.8\% execution accuracy. Even with subtask annotations as oracle hints, accuracy rises only to 30.1\%, demonstrating that these subtasks constitute fundamental bottlenecks.
By providing both a challenging, representative dataset and the framework for fine‑grained diagnostic evaluation, we believe \dataset{} offers a foundation for the next generation of robust, enterprise‑ready SQL agents.




\bibliographystyle{ACM-Reference-Format}
\bibliography{references}


\end{document}